\pdfoutput=1

\documentclass[11pt]{article}

\usepackage[preprint]{acl}

\usepackage{times}
\usepackage{latexsym}

\usepackage[T1]{fontenc}

\usepackage[utf8]{inputenc}

\usepackage{microtype}

\usepackage{inconsolata}

\usepackage{graphicx}

\usepackage{amsmath,amsfonts,bm}

\def\tabref#1{table~\ref{#1}}

\def\figref#1{figure~\ref{#1}}

\def\Secref#1{Section~\ref{#1}}

\def\eqref#1{equation~\ref{#1}}

\def\Algref#1{Algorithm~\ref{#1}}

\def\1{\bm{1}}

\DeclareMathAlphabet{\mathsfit}{\encodingdefault}{\sfdefault}{m}{sl}
\SetMathAlphabet{\mathsfit}{bold}{\encodingdefault}{\sfdefault}{bx}{n}

\def\gD{{\mathcal{D}}}
\def\gE{{\mathcal{E}}}

\def\gG{{\mathcal{G}}}

\def\gM{{\mathcal{M}}}

\def\gP{{\mathcal{P}}}

\def\gR{{\mathcal{R}}}
\def\gS{{\mathcal{S}}}

\newcommand{\E}{\mathbb{E}}

\usepackage{hyperref}
\usepackage{url}
\usepackage{paralist}
\usepackage{mathtools,amsthm,amssymb}
\usepackage{multirow}
\usepackage{booktabs}
\usepackage{xspace}
\usepackage{pifont}
\usepackage{tabularx}
\usepackage{caption}
\usepackage{array}
\newcolumntype{C}[1]{>{\centering\arraybackslash}m{#1}}
\newcolumntype{L}[1]{>{\arraybackslash}m{#1}}

\usepackage{algorithm}
\usepackage{algpseudocode}
\usepackage{setspace}
\usepackage{xcolor,colortbl}
\usepackage{enumitem}
\usepackage{wrapfig}
\usepackage{makecell}

\usepackage{pgfplots}
\usepackage{adjustbox}
\pgfplotsset{compat=1.18}

\definecolor{model0base}{RGB}{31,119,180} \definecolor{model0imp}{RGB}{143,187,218}
\definecolor{model1base}{RGB}{255,127,14} \definecolor{model1imp}{RGB}{255,191,135}
\definecolor{model2base}{RGB}{60,200,160} \definecolor{model2imp}{RGB}{158,228,208}
\definecolor{model3base}{RGB}{214,39,40} \definecolor{model3imp}{RGB}{235,147,148}
\definecolor{model4base}{RGB}{148,103,189} \definecolor{model4imp}{RGB}{202,179,222}
\definecolor{model5base}{RGB}{140,86,75} \definecolor{model5imp}{RGB}{198,171,165}
\definecolor{avgbase}{RGB}{128,128,128} \definecolor{avgimp}{RGB}{191,191,191}

\makeatletter
\algrenewcommand\ALG@beginalgorithmic{\linespread{1.2}\selectfont} 
\makeatother

\newcommand{\ours}{\textsc{Play2Prompt}\xspace}
\newcommand{\oursabbrev}{\textsc{P2P}\xspace}

\algrenewcommand\algorithmicrequire{\textbf{Input:}}
\algrenewcommand\algorithmicensure{\textbf{Output:}}
\let\Algorithm\algorithm
\renewcommand\algorithm[1][]{\Algorithm[#1]\setstretch{1.25}}

\newcommand{\e}[1]{{\small $#1$}}

\title{\ours: Zero-shot Tool Instruction Optimization\\ for LLM Agents via Tool Play}

\author{
  \textbf{Wei Fang\textsuperscript{$\dagger$}},
  \textbf{Yang Zhang\textsuperscript{$\ddagger$}},
  \textbf{Kaizhi Qian\textsuperscript{$\ddagger$}},
  \textbf{James Glass\textsuperscript{$\dagger$}},
  \textbf{Yada Zhu\textsuperscript{$\ddagger$}}
\\
  \textsuperscript{$\dagger$}Massachusetts Institute of Technology, Cambridge MA, USA \\
  \textsuperscript{$\ddagger$}MIT-IBM Watson AI Lab, Cambridge MA, USA
\\
\texttt{\hypersetup{urlcolor=black}\{\href{mailto:weifang@mit.edu}{weifang},glass\}@mit.edu},~\texttt{\{yang.zhang2,kqian\}@ibm.com},~\texttt{yzhu@us.ibm.com}
}

\begin{document}
\maketitle

\begin{abstract}
Large language models (LLMs) are increasingly integrated with specialized external tools, yet many tasks demand zero-shot tool usage with minimal or noisy documentation. 
Existing solutions rely on manual rewriting or labeled data for validation, making them inapplicable in true zero-shot settings. 
To address these challenges, we propose \ours, an automated framework that systematically ``plays'' with each tool to explore its input-output behaviors. 
Through this iterative trial-and-error process, \ours refines tool documentation and generates usage examples without any labeled data. 
These examples not only guide LLM inference but also serve as validation to further enhance tool utilization. 
Extensive experiments on real-world tasks demonstrate that \ours significantly improves zero-shot tool performance across both open and closed models, offering a scalable and effective solution for domain-specific tool integration\footnote{Source code is available at \texttt{\url{https://github.com/wfangtw/play2prompt}}.}.
\end{abstract} 

\section{Introduction}

Recently, there has been growing research interest in agentic large language model (LLM) frameworks, where, rather than having LLMs answer requests and queries from their own knowledge, LLMs can call a set of external tools with specialized capabilities. This allows LLMs to address more complex tasks and produce responses more accurately~\citep{mialon2023augmented,qin2024toollearningfoundationmodels}. One key challenge in developing agentic LLM frameworks is how to \textit{dynamically learn to use new, user-defined tools}, which is crucial because having a fixed, general-purpose tool set is often insufficient for real-world scenarios requiring domain-specific functionalities.

The existing mainstream paradigm for dynamically incorporating new tools is by supplementing user-defined tools at inference time in a zero- or few-shot manner via prompting~\citep{lu2023chameleon,shen2023hugginggpt}, leveraging zero-shot tool-calling capabilities of current LLMs that have been tuned with tool-use instructions. The success of this paradigm depends on providing sufficient information about the new tool in the prompt. Specifically, existing approaches generally rely on two types of information: \ding{182} \emph{Comprehensive tool documentation} detailing the tool’s functionalities and input/output formats, and \ding{183} \emph{In-context demonstrations} that include example queries and corresponding tool calls~\citep{hsieh2023tool, patil2023gorilla}.
Inadequate documentation can lead to failures in tool usage, such as syntax errors in both zero-shot and fine-tuned models~\citep{zhang2023syntax}, hallucinations due to incomplete or incorrect tool documentation~\citep{hsieh2023tool}, and diminished performance resulting from inadequate demonstrations~\citep{xu2023tool}. 

However, in many practical scenarios, it is often not realistic to rely on users, who are often non-experts in AI, to provide adequate documentation for their tools, nor to craft tool-use examples. When users do provide documentation, it may lack crucial details needed for LLMs to call the tools correctly. While automatic prompt optimization techniques~\citep{wang2024promptagent} could enhance tool documentation, they still require sufficient tool-use examples, which are unavailable in true zero-shot settings. In short, without tool-use examples, neither polished documentation nor in-context demonstrations can be supplied, leading to significant performance degradation in integrating new tools.

\begin{figure*}[t!]
    \centering
    \includegraphics[width=0.8\linewidth]{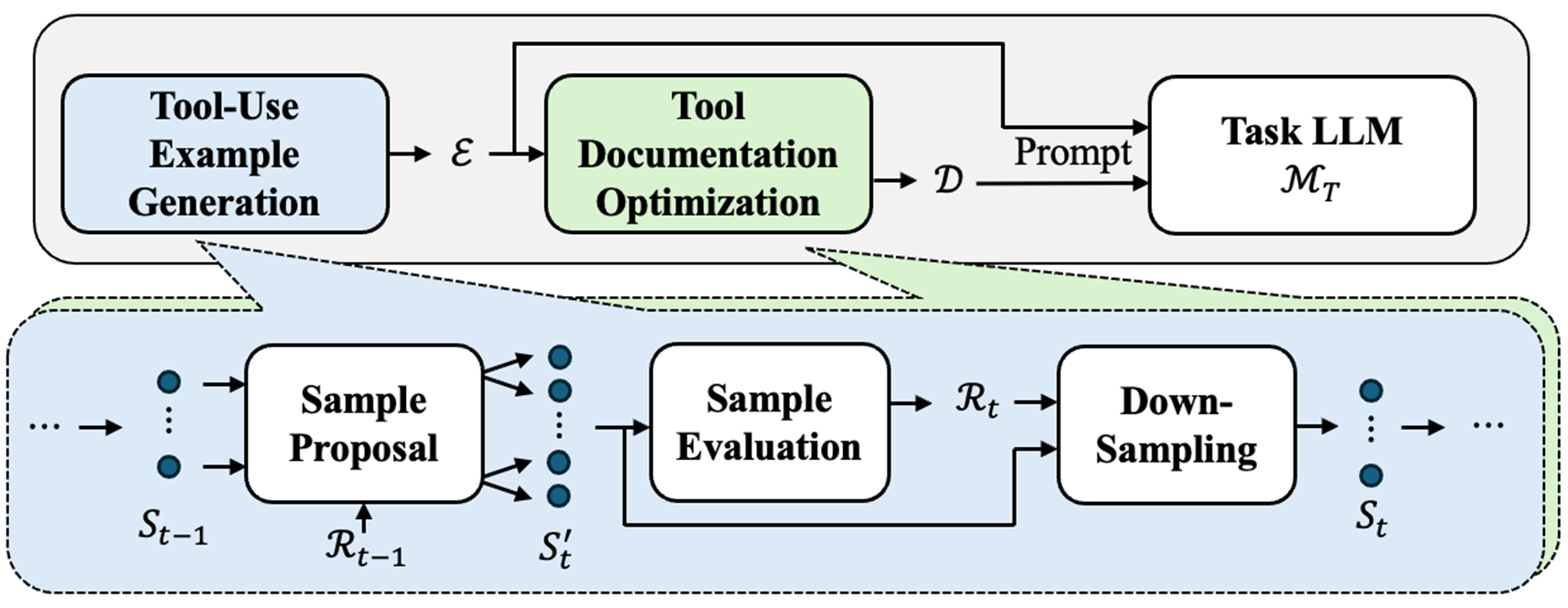}
    \caption{The \ours framework: Beam search iteratively searches tool-use examples, incorporating tool play into the proposal process. After examples are generated, beam search is once again applied to optimize documentation by incorporating tool-use outputs and errors. Finally, optimized tool-use examples and documentation are used as prompts for \e{\gM_T} at inference.}
    \label{fig:framework}
    \vspace{-1em}
\end{figure*}

To address these challenges in zero-shot tool utilization, we introduce \ours, an automated framework that generates both high-quality tool documentation and tool-use demonstrations, as illustrated in \figref{fig:framework}. 
Unlike prior works, \ours does not rely on any external labeled examples. 
Instead, it systematically interacts with the new tools---mimicking human trial-and-error---and observes both successful and failed attempts to gather evidence about each tool's correct usage.
Using insights from this ``tool-play'', \ours creates example demonstrations and refines the tool documentation that better guide LLMs in subsequent inference.

\ours consists of two steps. In Step 1, a set of tool-use examples are generated via a trial-and-error process, where an LLM agent iteratively call the target tool with different invocation parameters until correct invocation is found. Then, for each correct tool invocation instance, a query is generated such that it can be answered by the tool invocation, forming a question-answer pair as a tool-use example. In step 2, the tool documentation are refined, using the generated tool-use examples as a validation set. In both steps, We employ self-reflection~\citep{madaan2023self,pryzant-etal-2023-automatic,shinn2023reflexion} to generate error feedback, thereby directing the search algorithm towards progressively improved outputs. Because \ours operates entirely in a zero-shot manner and is inherently task-agnostic, it offers a practical and scalable solution for enhancing LLM tool utilization without additional labeled data or human intervention.

We evaluate \ours with benchmark on real-world scenarios. On the Berkeley Function-Calling Leaderboard~\citep{berkeley-function-calling-leaderboard} and the StableToolBench benchmark~\citep{guo-etal-2024-stabletoolbench}, our approach consistently surpasses baseline methods for both open~\citep{dubey2024llama,liu2025toolace,lin2025robust} and closed models~\citep{achiam2023gpt}.
Extensive experiments and analyses further underscore the robustness of our approach.

\noindent Our contributions can be summarized as follows:
\begin{itemize}[left = 0pt, noitemsep,topsep=0pt,parsep=1pt,partopsep=1pt] 
\item We propose \ours, a novel automated framework that iteratively refines tool documentation and generates usage examples, enabling more effective zero-shot tool utilization without any labeled data.
\item \ours integrates a search-based trial-and-error process augmented with self-reflection, allowing LLMs to explore and ``play'' with tools to refine both tool documentation and demonstrations for enhanced performance. 
\item \ours is entirely zero-shot, 
 scalable, and task-agnostic, making it broadly applicable across diverse tools and domains, and practical for large-scale enhancement of LLM tool use without additional manual effort. 
\end{itemize}

\section{Methodology}

\subsection{Problem Formulation and Notation.} 
\label{sec:notation}
A typical agentic LLM framework contains two components: \ding{182} A task LLM, denoted as \e{\mathcal{M}_T}, and \ding{183} a set of tools, \e{\mathcal{F} = \{F_{1:K}\}}, where \e{K} represents the number of tools, and \e{1:K} represents a set of indices running from \e{1} to \e{K}. Given an input query \e{x}, rather than directly answering it based on its own knowledge, the task LLM \e{\mathcal{M}_T} first selects a sequence of \e{N} tools, \e{(F_{k_{1:N}})}, generates the appropriate input parameters, \e{I_{k_n}}, to call each tool, \emph{i.e.}, \e{F_{k_n}(I_{k_n})}, and finally produces the answer \e{y} based on the tool outputs.

We consider the setting where the tool list \e{\mathcal{F}} is ad-hoc and dynamic. In order for the task LLM to learn to use the tools in \e{\mathcal{F}} at inference time, we follow the conventional prompting-based pipeline~\citep{lu2023chameleon,hsieh2023tool}, where the prompt, in addition to an instruction, contains the following tool-specific information:

\noindent\e{\bullet} \textbf{Tool Documentation}, denoted as \e{\mathcal{D}=\{D_{1:K}\}};

\noindent\e{\bullet} \textbf{In-context examples of tool-use}, denoted as \e{\mathcal{E} = \{E_{1:M}\}}, where each example \e{E_m} contains an input query, \e{x}, the tool invocation details, \e{(F_{k_{1:N}}, I_{k_{1:N}})}, and the answer, \e{y}.

In many real-world scenarios where users supply their own tool list, it is unrealistic to require them to supply high-quality documentation or tool-use examples. To model these scenarios, we adopt a challenging zero-shot setting with the following constraints: \ding{182} The initial documentation \e{\mathcal{D}_0} is sub-optimal and may lack important details, and \ding{183} No tool-use examples \e{\mathcal{E}} are available. Given these constraints, our objective is to generate tool-use examples and refine the tool documentation to enhance the task LLM’s tool-use performance.

\subsection{\ours Overview}

The primary goal of \ours is to utilize knowledge gained from tool interactions to optimize tool documentation and example demonstrations.
\ours consists of the following two steps. \textit{First}, a tool-use example set, \e{\mathcal{E}}, is generated. \textit{Second}, using \e{\mathcal{E}} as the validation set, a refined tool documentation \e{\mathcal{D}} is generated based on the initial one \e{\mathcal{D}_0}. After the two steps are accomplished, \e{\mathcal{D}} and a subset of \e{\mathcal{E}} will be fed as the prompt to the task LLM during inference.

In both steps, we adopt a tree search framework similar to \citet{wang2024promptagent} to generate tool-use examples or tool documentation. Therefore, we will first briefly introduce the beam search framework (see figure~\ref{fig:framework}) in Section~\ref{subsec:beamsearch}, and then elaborate on the two steps in Sections~\ref{subsec:example_gen} and \ref{subsec:doc_gen}, respectively.

\subsection{Search Framework}
\label{subsec:beamsearch}

The beam search framework is an iterative algorithm to generate high-quality samples, denoted \e{\mathcal{S}=\{S^{(i)}\}}, which correspond to \e{\mathcal{E}} in step 1, and \e{\mathcal{D}} in step 2. At the start of iteration \e{t}, the algorithm has access to \e{\mathcal{S}_{t-1} = \{S^{(i)}_{t-1}\}}, which are samples generated in the previous iteration, as well as their corresponding reward \e{\{R_{t-1}^{(i)}\}}, which depicts the quality of each sample. Then iteration \e{t} involves the following procedure to generate a set of improved samples:

\noindent \e{\bullet} \textbf{Sample Proposal.} For each old sample, \e{S^{(i)}_{t-1}}, generate \e{L} new samples from the proposal distribution \e{p(S^{(i)}_t | S^{(i)}_{t-1}, R^{(i)}_{t-1})}, based on the reward of the old sample. This is accomplished by prompting a generator LLM, denoted as \e{\mathcal{G}}, to perform self-reflection on why the old sample is imperfect, how to improve the sample quality, and finally generate the improved samples. All the new samples form a new sample set, denoted as \e{\mathcal{S}'_t}, whose size is \e{L} times the size of \e{\mathcal{S}_{t-1}}.

\noindent \e{\bullet} \textbf{Sample Evaluation.} For each new sample \e{S_t^{(i)}} in \e{\mathcal{S}'_t}, compute its reward \e{R_t^{(i)}}.

\noindent \e{\bullet} \textbf{Subsampling.} Trim \e{\mathcal{S}'_t} down to the size of \e{\mathcal{S}_{t-1}} by keeping the samples with the highest reward. Denote the trimmed sample set as \e{\mathcal{S}_t}.

The iterations terminate when the pre-set maximum number of iterations is reached. With the beam search framework, the algorithm design boils down to designing \ding{182} the sample proposal distribution, and \ding{183} the reward and feedback of each sample. In the following sections, we detail how these design choices are set in tool-use example generation and tool documentation optimization.

\subsection{Tool-Use Example Generation}
\label{subsec:example_gen}

In this work, we only consider generating examples where only a single tool is used. We will show that (\Secref{sec:exp_main}) the task LLM can still learn to solve queries that require multiple tools with the examples of single tool use. In this way, examples for different tools can be generated separately. Specifically, examples of using tool \e{F_k} take the form of \e{E=(x, F_k, I_k, y)}, which can be generated via the beam search framework in Section~\ref{subsec:beamsearch}, with the design choices detailed below.

\paragraph{Sample Proposal.} The sample proposal is performed by prompting an example generator LLM, denoted as \e{\mathcal{G}_E}. Since the tool is fixed to \e{F_k}, it only needs to propose new samples for the query \e{x}, the invocation parameters \e{I_k}, and the final answer \e{y}. However, the challenge lies in the limited information available about \e{F_k}—only an (incomplete) initial documentation \e{\mathcal{D}_0} and no user-supplied examples—making it likely that the generated samples fail to invoke the tool correctly.

To address this, we generate samples in \textbf{reverse order}: first, we explore valid invocations \e{I_k}, observe the tool’s output, and then construct a corresponding query \e{x} and answer \e{y}. This approach effectively \textit{"plays with" the tool} to understand its behavior before defining its use cases.

Legitimate samples of \e{I_k} are generated in a rejection-sampling process, where \e{\mathcal{G}_E} first generates a tentative sample invocation given \e{\mathcal{D}_0}, observes the tool outputs and error messages, performs self-reflection, and generates the next one. Note that this inner loop is nested in the outer loop of the beam search framework, so the rejection sampling is also conditional on the tool-use examples generated in the previous outer iteration \e{t-1}, along with their reward (see Section~\ref{subsec:beamsearch}), except for in outer iteration 0. This inner loop terminates at a fixed number of steps, after which \e{L} legitimate invocations are selected.\footnote{If the number of legitimate invocations is smaller than \e{L}, we will limit the number of proposed samples accordingly.}

For each sampled legitimate invocation, we feed the invocation and the tool output to \e{\mathcal{G}_E}, which is then prompted to generate a query and an answer by performing \e{N_E} steps of self-reflecting refinement, which concludes the sample proposal procedure.

\paragraph{Reward Design.} For each generated sample \e{E^{(i)}=(x^{(i)}, F_k, I_k^{(i)}, y^{(i)})}, the corresponding reward consists of two terms, 
\begin{equation}
\small
R^{(i)} = R^{(i)}_q + \lambda R^{(i)}_e.
\label{eq:example_reward}
\end{equation}
\e{R^{(i)}_q} evaluates the quality of the generated example, including clarity and coherence between the input query and tool use. This is evaluated by prompting \e{\mathcal{G}_E} to output a score of 1-3 given the grading criteria. \e{R^{(i)}_e} evaluates the performance of the task LLM \e{\mathcal{M}_T} in answering the query in this example:
\begin{equation}
\small
    R^{(i)}_e = -\mathcal{P}\{\mathcal{M}_T(x^{(i)}; \mathcal{D}_0, \varnothing); y^{(i)}, F_k, I_k^{(i)}\},
    \label{eq:reward_eval}
\end{equation}
where \e{\mathcal{P}\{\hat{M}; y, F_k, I_k, \}} denotes the task performance metric (the higher the better, \emph{e.g.}, accuracy) of the model output \e{\hat{M}} against the ground-truth answer \e{y} and tool invocation \e{F_k, I_k}; \e{\mathcal{M}_T(x^{(i)}; \mathcal{D}_0, \varnothing)} denotes the output of the task LLM in answering the query \e{x^{(i)}} given the initial documentation \e{\mathcal{D}_0} and \emph{no in-context examples} (because we don't have any yet at this stage).

The negative sign in Eq.~\ref{eq:reward_eval} indicates that we encourage \textit{difficult examples} -- examples that the task LLM cannot get right, because difficult examples bring more surprise to the task LLM and thus are more effective in shaping the LLM's behavior.

\subsection{Tool Documentation Optimization}
\label{subsec:doc_gen}

The goal of tool documentation optimization is to generate improved tool documentation \e{\mathcal{D}} based on the initial one \e{\mathcal{D}_0}, which is again achieved by the beam search framework, where the documentation sample with the highest reward will be chosen as the final tool documentation.

\paragraph{Sample Proposal.} The sample proposal process primarily follows the approach described in Section~\ref{subsec:beamsearch}. We provide the generator LLM \e{\mathcal{G}_D}, which differs from \e{\mathcal{G}_E} used in the previous step, with the current documentation \e{\gD^{(i)}} along with tool use errors from \e{\gM_T}. By conditioning the proposal distribution on these errors, we inform \e{\mathcal{G}_D} of the documentation’s deficiencies or ambiguities, prompting more effective revisions.

\paragraph{Reward Design.} The reward is computed on the tool-use example set \e{\mathcal{E}} generated in the previous step, essentially treating \e{\mathcal{E}} as the validation set for documentation optimization. For each tool documentation sample, \e{\mathcal{D}^{(i)}}, the corresponding reward is the tool use performance given  \e{\mathcal{D}^{(i)}} on a small batch in the validation set:
\begin{equation}
    \small
    R^{(i)} = \mathbb{E}_{(x, F, I, y) \in \mathcal{E}} \big[ \mathcal{P}\{\mathcal{M}_T(x; \mathcal{D}^{(i)}, \varnothing); y, F, I\} \big].
    \label{eq:reward_demo}
\end{equation}
By comparing Eqs.~\ref{eq:reward_eval} and \ref{eq:reward_demo}, it can be observed that tool-use example generation and tool documentation optimization have adversarial objectives, the former seeking to reduce the tool use performance (while maintaining quality and alignment), and the latter to improve it. This resembles the active learning strategy of choosing high-loss examples~\citep{Yoo_2019_CVPR}, which reveals that maximizing the performance on the most difficult examples leads to high learning efficiency.

\section{Experiments}
\subsection{Experimental Setup}\label{sec:exp_setup}
\paragraph{Dataset: Berkeley Function-Calling Leaderboard (BFCL).}
We evaluate on the Berkeley Function-Calling Leaderboard~\citep{berkeley-function-calling-leaderboard}, a benchmark of real-world data that assesses LLMs' tool-use abilities.
Its data includes a non-executable subset evaluated by abstract syntax trees and an executable subset assessed by running the functions.
We use the executable subset (referred to as Executable), because actual \emph{tool-play} is central to our approach, reflecting realistic scenarios in which most real-world APIs are callable.
This subset has four categories of Python functions (single-tool, multi-tool, parallel tool-calling, and multi-tool parallel tool-calling) and plus one category of REST functions, yielding  310 test queries.
\paragraph{Dataset: StableToolBench.} We also evaluate on StableToolBench~\citep{guo-etal-2024-stabletoolbench}, an updated version of ToolBench~\citep{qin2024toolllm}, one of the most widely-used tool-use dataset, which addressed RapidAPIs' instability through a fallback system with caching and API simulation. 
Our experiments use all six of its testing subsets, including single-tool (I1), multi-tool (I2-same category and I3-different category) queries. 
Individual APIs, grouped under ``tools'', correspond to $F_k$, so I1 test queries usually require calls to multiple APIs within a tool and are not single-tool under our definition.
The original dataset's categorization based on tool overlap with training data are thus less relevant for our strictly zero-shot setting.
In total, these six subsets cover 790 queries, spanning 2479 unique APIs with an average of $5.3$ APIs per query.
\begingroup
\setlength{\tabcolsep}{4pt}
\begin{table*}[t!]
\small
\centering
\begin{tabular}{lL{3.4cm}C{1.2cm}C{1.2cm}C{1.2cm}C{1.2cm}C{1.2cm}>{\columncolor[gray]{0.9}}C{1.2cm} >{\columncolor[gray]{0.9}}C{1.2cm}}
\toprule
\bf Base Model & \bf Method  & \bf Simple-Python & \bf Simple-REST & \bf Multiple & \bf Parallel & \bf Multiple-Parallel & \bf Weighted Avg & \bf Avg \\
\midrule
\multirow{2}{*}{LLaMA-8B} & Prompting & 96.0 & 70.0 & 96.0 & 90.0 & 77.5 & 86.6 &  85.9 \\
 &+\ours & 97.0 & 87.1 & 96.0 & 92.0 & 92.5 & \bf 93.1 & \bf 92.9 \\
\midrule
\multirow{2}{*}{LLaMA-70B} & Prompting & 100.0 & 91.4 & 96.0 & 84.0 & 82.5 & 89.6 & 90.8 \\
&+\ours  & 100.0 & 91.4 & 98.0  & 88.0 & 95.0 & \bf 94.2 & \bf 94.5  \\
\midrule
\multirow{2}{*}{GPT-3.5} & Function-calling & 97.0 & 94.3 & 90.0 & 86.0 & 67.5 & 84.8 & 87.0 \\
&+\ours  & 98.0 & 95.7 & 94.0 & 90.0 & 85.0 & \bf 91.5 & \bf 92.5 \\
\midrule
\multirow{2}{*}{GPT-4o} & Function-calling & 98.0 & 98.6 & 94.0 & 94.0 & 77.5 & 91.0 & 92.4 \\
&+\ours  & 99.0 & 95.7 & 98.0 & 92.0 & 90.0 & \bf 94.3 & \bf 94.9 \\
\midrule
\multirow{2}{*}{ToolACE-8B} & Function-calling & 96.0 & 92.9 & 92.0 & 86.0 & 70.0 & 85.6 & 87.4 \\
&+\ours  & 95.0 & 90.0 & 94.0 & 90.0 & 82.5 & \bf 89.8 & \bf 90.3 \\
\midrule
\multirow{2}{*}{Hammer-7B} & Function-calling & 96.0 & 72.9 & 88.0 & 86.0 & 70.0 & 82.1 & 82.6 \\
&+\ours  & 95.0 & 82.8 & 92.0 & 86.0 & 80.0 & \bf 86.7 & \bf 87.2 \\
\bottomrule
\end{tabular}
\caption{Results on BFCL Executable. Accuracy scores are shown.}
\label{tab:bfcl_main}
\end{table*}
\endgroup

\begingroup
\setlength{\tabcolsep}{4pt}
\begin{table*}[t!]
\small
\centering
\begin{tabular}{lL{3.4cm}C{1.2cm}C{1.2cm}C{1.2cm}C{1.2cm}C{1.2cm}C{1.2cm}>{\columncolor[gray]{0.9}}C{1.2cm}}
\toprule
\bf Base Model & \bf Method  & \bf I1-Inst & \bf I1-Cat & \bf I1-Tool & \bf I2-Inst & \bf I2-Cat & \bf I3-Inst & \bf Avg \\ \midrule
\multirow{4}{*}{LLaMA-8B} & ReAct & 50.6 & 59.3 & 53.2 & 58.0 & 61.3 & 52.3 & 55.8 \\
 &ReAct+EasyTool & 54.7 & 58.9 & 57.9 & 56.1  & 64.2 & 48.1  & 56.7 \\
 &ReAct+DRAFT & - & - & - & -  & 62.6 & 57.9  & - \\
 &ReAct+\ours & 56.6 & 65.7 &  60.9 &  62.5 &  63.5 &  63.8 &  \bf 62.2 \\
\midrule
\multirow{4}{*}{LLaMA-70B} & ReAct & 58.4 & 68.3 & 61.1 & 63.1 & 63.2 & 64.3 & 63.1 \\
 &ReAct+EasyTool & 59.0 & 70.3 & 63.3  & 67.6  & 70.6  & 63.0 & 65.6 \\
 &ReAct+DRAFT & - & - & -  & -  & 68.5  & 62.3 & - \\
&ReAct+\ours  &  67.5 &  73.6 &  64.1 &  66.5 &  72.8 &  70.7 &  \bf 69.2 \\
\midrule
\multirow{4}{*}{GPT-3.5} & ReAct & 56.0 & 64.6 & 67.4 & 56.0 & 63.4 & 57.4 &  60.7 \\
 &ReAct+EasyTool & 57.3 & 61.4 & 69.5 & 61.5  & 68.3  & 60.7  & 63.1 \\
 &ReAct+DRAFT & - & - & - & -  & 68.3  & 56.1  & - \\
&ReAct+\ours  &  61.1 &  66.6 & 66.2 &  67.0 & 68.3 &  66.3 & \bf 65.9 \\
\midrule
\multirow{4}{*}{GPT-4o} & ReAct & 54.0 & 69.7 & 66.1 & 59.8 & 65.3 & 61.7 &  62.8 \\
 &ReAct+EasyTool & 49.9 & 69.0 & 65.2 & 61.3  & 65.3  & 52.5  & 60.5 \\
 &ReAct+DRAFT & - & - & - & -  & 62.1  & 63.9  & - \\
&ReAct+\ours  & 60.7  & 68.7 & 71.9 & 58.8 & 63.7 & 65.6 & \bf 64.9 \\
\bottomrule
\end{tabular}
\caption{Results on StableToolBench. Solvable pass rates are shown.}
\vspace{-1em}
\label{tab:stb_main}

\end{table*}
\endgroup
\paragraph{Inference and Evaluation.} 
We adhere to official inference settings of each benchmark, where a set of tools is provided for each test query. 
For task LLM \e{\gM_T}, we tested the 8B and 70B LLaMA models~\citep{dubey2024llama}, and GPT-3.5 and GPT-4o were used for GPT~\citep{achiam2023gpt}. 
We also tested on BFCL two state-of-the-art LLMs trained specifically for tool-calling, namely ToolACE~\citep{liu2025toolace} and Hammer~\citep{lin2025robust}, with both models achieving high rankings on the BFCL leaderboard as of this publication.
For BFCL, single-turn prompting with official prompts is used as the baseline inference method for LLaMA models, while direct function-calling mode is used for the GPT models, ToolACE, and Hammer.
We do not provide tool-use examples or additional documentation in these baseline runs, complying with a zero-shot setting.
For \ours, optimized in-context examples and documentation are supplied as prompts and runs the baseline inference methods.
It supports multi-tool queries by independently optimizing examples and documentation for each tool and then performing inference.  
We use the official evaluation metric of accuracy, with exact or structural matches depending on categories to determine correctness.
We report the average across five categories, and, following the official setup, the weighted average (``Simple'' categories weighted by $0.5$).

StableToolBench employs a more complex chain-of-thought inference method, ReAct~\citep{yao2023react}. 
We compare against EasyTool~\cite{yuan2024easytool}, which uses direct prompting to optimize tool documentation and generate usage scenarios, but requires documentation and labeled in-context examples from additional tools and is thus not entirely zero-shot\footnote{We use the optimized tool documentation and usage scenarios provided at \texttt{\url{https://github.com/microsoft/JARVIS/tree/main/easytool}}. The inference setting of \citet{yuan2024easytool} differs from ours as we follow the official inference prompts provided with StableToolBench.}.
Additionally, we compare against DRAFT~\citep{quexploration}, a concurrent work, which optimizes tool documentation only in an iterative manner~\footnote{We use the optimized documentation very recently open-sourced at~\url{https://github.com/quchangle1/DRAFT}. Note that ~\citet{quexploration} tested only on the I2-Cat and I3-Inst subsets.}. 
An evaluation LLM is used to judge whether a response adequately answers a user query, due the dataset's free-form output design. 
We follow the official pipeline, using official ReAct prompts and report solvable pass rate, which measures the percentage of queries deemed solvable by the evaluation LLM. 
Further details are provided in appendix~\ref{appx:inference}.
\paragraph{Optimization Details.} 
\ours first uses beam search to optimize tool-use examples. 
We set \e{\lambda=1}, \e{N_E=3}, limit depth to \e{3}, and explore \e{L=3} beams per node. We use beam width \e{W=10} for BFCL and \e{W=3} for StableToolBench to generate and select the top \e{W} examples for each tool, which then are passed to the documentation optimization procedure. 
Beam search is again applied to select the best tool documentation.
We employ \texttt{llama-3.1-8b-instruct} as both \e{\gG_E} and \e{\gG_D}. 

\begin{figure*}
\centering
\begin{adjustbox}{max width=\textwidth, center}
\begin{tikzpicture}
\begin{axis}[
    bar width=0.92,
    width=28cm,
    height=10cm,   
    ylabel={Accuracy (\%)},
    ylabel style={font=\large},
    yticklabel style={font=\normalsize},
    ymin=60, 
    ymax=98,
    ytick={65,75,85,95},
    ymajorgrids=true,
    xmajorgrids=false,
    grid style={dashed, black!80}, 
    xtick={3, 12, 21}, 
    xticklabels={$p = 0.0$, $p = 0.5$, $p = 1.0$},
    xticklabel style={font=\Large}, 
    xtick style={draw=none},
    area legend, 
    legend style={
        at={(0.5,1.08)}, 
        anchor=south,
        legend columns=-1,
        draw=none, fill=none,
        font=\Large, 
        column sep=1em, 
    },
    every axis plot/.append style={draw=black, very thin},
    enlarge y limits={upper, value=0.02}, 
    enlarge x limits={abs=0.75}, 
    clip=false
]

\tikzset{barval/.style={font=\footnotesize\bfseries, black, inner sep=1pt}}


\addplot[ybar, fill=model0imp, forget plot] coordinates {(0, {86.6+6.5}) (9, {81.3+4.8}) (18, {73.7+7.0})};
\addplot[ybar, fill=model0base] coordinates {(0, 86.6) (9, 81.3) (18, 73.7)}; 
\addlegendentry{LLaMA-8B}
    \node[barval] at (axis cs:0, 89.85) {+\pgfmathprintnumber[fixed,precision=1]{6.5}\%};
    \node[barval] at (axis cs:9, 83.7)  {+\pgfmathprintnumber[fixed,precision=1]{4.8}\%};
    \node[barval] at (axis cs:18, 77.2) {+\pgfmathprintnumber[fixed,precision=1]{7.0}\%};

\addplot[ybar, fill=model1imp, forget plot] coordinates {(1, {89.6+4.6}) (10, {84.4+4.4}) (19, {75.9+5.4})};
\addplot[ybar, fill=model1base] coordinates {(1, 89.6) (10, 84.4) (19, 75.9)}; 
\addlegendentry{LLaMA-70B}
    \node[barval] at (axis cs:1, 91.9)  {+\pgfmathprintnumber[fixed,precision=1]{4.6}\%};
    \node[barval] at (axis cs:10, 86.6) {+\pgfmathprintnumber[fixed,precision=1]{4.4}\%};
    \node[barval] at (axis cs:19, 78.6) {+\pgfmathprintnumber[fixed,precision=1]{5.4}\%};

\addplot[ybar, fill=model2imp, forget plot] coordinates {(2, {84.8+6.7}) (11, {74.9+10.9}) (20, {68.9+8.3})};
\addplot[ybar, fill=model2base] coordinates {(2, 84.8) (11, 74.9) (20, 68.9)}; 
\addlegendentry{GPT-3.5}
    \node[barval] at (axis cs:2, 88.15) {+\pgfmathprintnumber[fixed,precision=1]{6.7}\%};
    \node[barval] at (axis cs:11, 80.35) {+\pgfmathprintnumber[fixed,precision=1]{10.9}\%};
    \node[barval] at (axis cs:20, 73.05) {+\pgfmathprintnumber[fixed,precision=1]{8.3}\%};

\addplot[ybar, fill=model3imp, forget plot] coordinates {(3, {91.0+3.3}) (12, {87.3+3.0}) (21, {76.5+6.7})};
\addplot[ybar, fill=model3base] coordinates {(3, 91.0) (12, 87.3) (21, 76.5)}; 
\addlegendentry{GPT-4o}
    \node[barval] at (axis cs:3, 92.65)  {+\pgfmathprintnumber[fixed,precision=1]{3.3}\%};
    \node[barval] at (axis cs:12, 88.8)  {+\pgfmathprintnumber[fixed,precision=1]{3.0}\%};
    \node[barval] at (axis cs:21, 79.85) {+\pgfmathprintnumber[fixed,precision=1]{6.7}\%};

\addplot[ybar, fill=model4imp, forget plot] coordinates {(4, {85.6+4.2}) (13, {77.9+5.4}) (22, {76.1+6.0})};
\addplot[ybar, fill=model4base] coordinates {(4, 85.6) (13, 77.9) (22, 76.1)}; 
\addlegendentry{ToolACE-8B}
    \node[barval] at (axis cs:4, 87.7)  {+\pgfmathprintnumber[fixed,precision=1]{4.2}\%};
    \node[barval] at (axis cs:13, 80.6) {+\pgfmathprintnumber[fixed,precision=1]{5.4}\%};
    \node[barval] at (axis cs:22, 79.1) {+\pgfmathprintnumber[fixed,precision=1]{6.0}\%};

\addplot[ybar, fill=model5imp, forget plot] coordinates {(5, {82.1+4.6}) (14, {76.2+6.4}) (23, {73.2+9.4})};
\addplot[ybar, fill=model5base] coordinates {(5, 82.1) (14, 76.2) (23, 73.2)}; 
\addlegendentry{Hammer-7B}
    \node[barval] at (axis cs:5, 84.4)  {+\pgfmathprintnumber[fixed,precision=1]{4.6}\%};
    \node[barval] at (axis cs:14, 79.4) {+\pgfmathprintnumber[fixed,precision=1]{6.4}\%};
    \node[barval] at (axis cs:23, 77.9) {+\pgfmathprintnumber[fixed,precision=1]{9.4}\%};

\addplot[ybar, fill=avgimp, forget plot] coordinates {(6, {86.617+4.983}) (15, {80.333+5.817}) (24, {74.050+7.133})};
\addplot[ybar, fill=avgbase] coordinates {(6, 86.617) (15, 80.333) (24, 74.050)}; 
\addlegendentry{Average}
    \node[barval] at (axis cs:6, 89.1085)  {+\pgfmathprintnumber[fixed,precision=1]{4.983}\%};
    \node[barval] at (axis cs:15, 83.2415) {+\pgfmathprintnumber[fixed,precision=1]{5.817}\%};
    \node[barval] at (axis cs:24, 77.6165) {+\pgfmathprintnumber[fixed,precision=1]{7.133}\%};

\end{axis}
\end{tikzpicture}
\end{adjustbox}

\caption{Average (weighted) accuracy improvements with \ours on BFCL, for different parameter description dropout $p$.}
\label{fig:robustness}
\end{figure*}

\subsection{Results and Analyses}\label{sec:exp_main}
\paragraph{Results on BFCL and StableToolbench.}
In \tabref{tab:bfcl_main}, we summarize the results on BFCL with LLaMA and GPT task models, both with and without tool-use examples and tool documentation produced by \ours.
With \ours, absolute gains of 4-7\% are observed for the open models and GPT-3.5, while GPT-4o achieves 3\% increase in average accuracy despite its already high baseline.
Notably, \ours addresses challenging categories such as REST for LLaMA-8B and Hammer-7B, and Multiple-Parallel for all models, yielding improvements of 10-17\%.
These gains suggest that optimized in-context examples and documentation can correct specific shortcomings in tool usage.
Moreover, for the more difficult multi-tool queries in Multiple and Multiple-Parallel, we see larger gains compared to single-tool, even when we optimize each tool independently.

In \tabref{tab:stb_main}, we show the solvable pass rates on StableToolBench for LLaMA and GPT models with and without \ours.
Our approach surpasses all baselines with average gains of 5-7\% across LLaMA models and GPT-3.5, and outperforms the specifically designed EasyTool, which is not fully zero-shot as it requires in-context demonstrations during optimization.
\ours is also consistent across subsets, avoiding large performance drops in multiple subsets when using EasyTool, highlighting the benefits of real-time ``tool-play'' and evaluation of candidate examples and documentation during optimization.
Moreover, as GPT-4o exhibits more sensitivity to documentation and in-context demonstration changes on this dataset, EasyTool degrades GPT-4o on all subsets, whereas \ours achieves a 2\% improvement on average.
Overall, \ours essentially boosts performance of models up to the baseline performance of the larger models.
Compared to DRAFT, which optimizes tool documentation only, \ours outperforms for all four models on both I2-Cat and I3-Inst. 
This suggests that optimizing not only documentation, as DRAFT does, but also generating tool-use examples, can synergize and achieve even larger improvements. 
Additionally, our optimization model of LLaMA-8B is much smaller than the GPT-4o model used in DRAFT, further confirming the effectiveness of \ours.
It is noteworthy that most test samples for all 6 subsets in StableToolBench, as well as the Multiple and Multiple-Parallel subsets in BFCL, are multi-tool queries (see \Secref{sec:exp_setup}), underscoring the effectiveness of \ours, which operates on single tools independently during optimization.

\begingroup
\setlength{\tabcolsep}{4pt}
\begin{table}[t!]
\small
\centering
\begin{tabular}{lL{2.5cm}C{1.15cm}C{1.15cm}}
\toprule
\bf Base Model & \bf Method  & BFCL & STB  \\
\midrule
\multirow{4}{*}{LLaMA-8B} & Baseline & 85.9 & 55.8 \\
 &+\oursabbrev-Desc & 89.9  & 57.9 \\
 &+\oursabbrev-Demo & 90.8 & 59.5 \\
 &+\oursabbrev & \bf  92.9 & \bf 62.2 \\
\midrule
\multirow{4}{*}{LLaMA-70B} & Baseline & 90.8 & 63.1 \\
 &+\oursabbrev-Desc & 93.6  & 64.4 \\
 &+\oursabbrev-Demo & 92.7 & 67.8 \\
&+\oursabbrev  & \bf  94.5 & \bf 69.2 \\
\midrule
\multirow{4}{*}{GPT-3.5} & Baseline &  87.0 & 60.7 \\
 &+\oursabbrev-Desc & 89.0 & 63.5 \\
 &+\oursabbrev-Demo &  91.9 & 65.3 \\
 &+\oursabbrev  & \bf 92.5 & \bf 65.9 \\
\bottomrule
\end{tabular}
\caption{Ablation on \ours(\oursabbrev) using generated example demonstrations only (\oursabbrev-Demo) and generated descriptions only (\oursabbrev-Desc). Scores indicate average accuracy for BFCL and average solvable pass rate for StableToolBench (STB).}
\label{tab:exp_demo_desc}
\vspace{-1.2em}
\end{table}
\endgroup

\paragraph{Robustness of \ours.}
To evaluate the robustness of \ours under incomplete documentation, we simulate noisy tool descriptions on the BFCL Executable dataset by randomly dropping parameter descriptions with increasing probabilities $p \in \{0.0, 0.5, 1.0\}$, while retaining overall tool descriptions and parameter names. 
This setup reflects varying degrees of real-world documentation sparsity.
As expected, the degradation of documentation leads to reduced baseline accuracy across all models. 

Despite these challenges, \ours consistently improves tool-use performance across all models and noise levels. 
As shown in \figref{fig:robustness}, \ours achieves steady accuracy gains, with average improvements growing from 5.0\% at $p=0.0$ and reaching 7.1\% at $p=1.0$. 
Notably, the benefit of \ours becomes even more pronounced as documentation quality deteriorates, demonstrating its ability to compensate for missing parameter details and its strong robustness in low-resource settings.
Detailed per-model results can be found in appendix~\ref{appx:robustness}.

\paragraph{Comparing Demonstrations and Documentation.}
We investigate how much the optimized examples contribute to \ours's performance gains compared documentation.
Using the same optimization procedure, we test two inference settings: one employs new in-context demonstrations alongside original documentation, and the other uses updated documentation without demonstrations.
As shown in \tabref{tab:exp_demo_desc}, employing only optimized demonstrations yields larger improvements than solely relying on optimized documentation, and combining both consistently achieves the best results.
These findings suggest that demonstrations and documentation complement each other in guiding LLMs' tool use, confirming the effectiveness of our two-stage approach.

\begingroup
\setlength{\tabcolsep}{4pt}
\begin{table}[t!]
\small
\centering
\begin{tabular}{L{1.7cm}L{3.8cm}C{1.15cm}}
\toprule
\bf Base Model & \bf Method  & \bf STB  \\
\midrule
\multirow{3}{*}{LLaMA-8B} & ReAct & 55.8 \\
 &+\ours-Easy & 60.7 \\
 &+\ours & \bf 62.2 \\
\midrule
\multirow{3}{*}{LLaMA-70B} & ReAct & 63.1 \\
 &+\ours-Easy & 68.2 \\
&+\ours & \bf 69.2 \\
\bottomrule
\end{tabular}
\caption{Ablation on demonstration difficulty for \ours. We report average solvable pass rates.}
\label{tab:exp_hard}
\vspace{-1em}
\end{table}
\endgroup

\begin{figure*}[ht!]
    \centering
    \includegraphics[width=\linewidth]{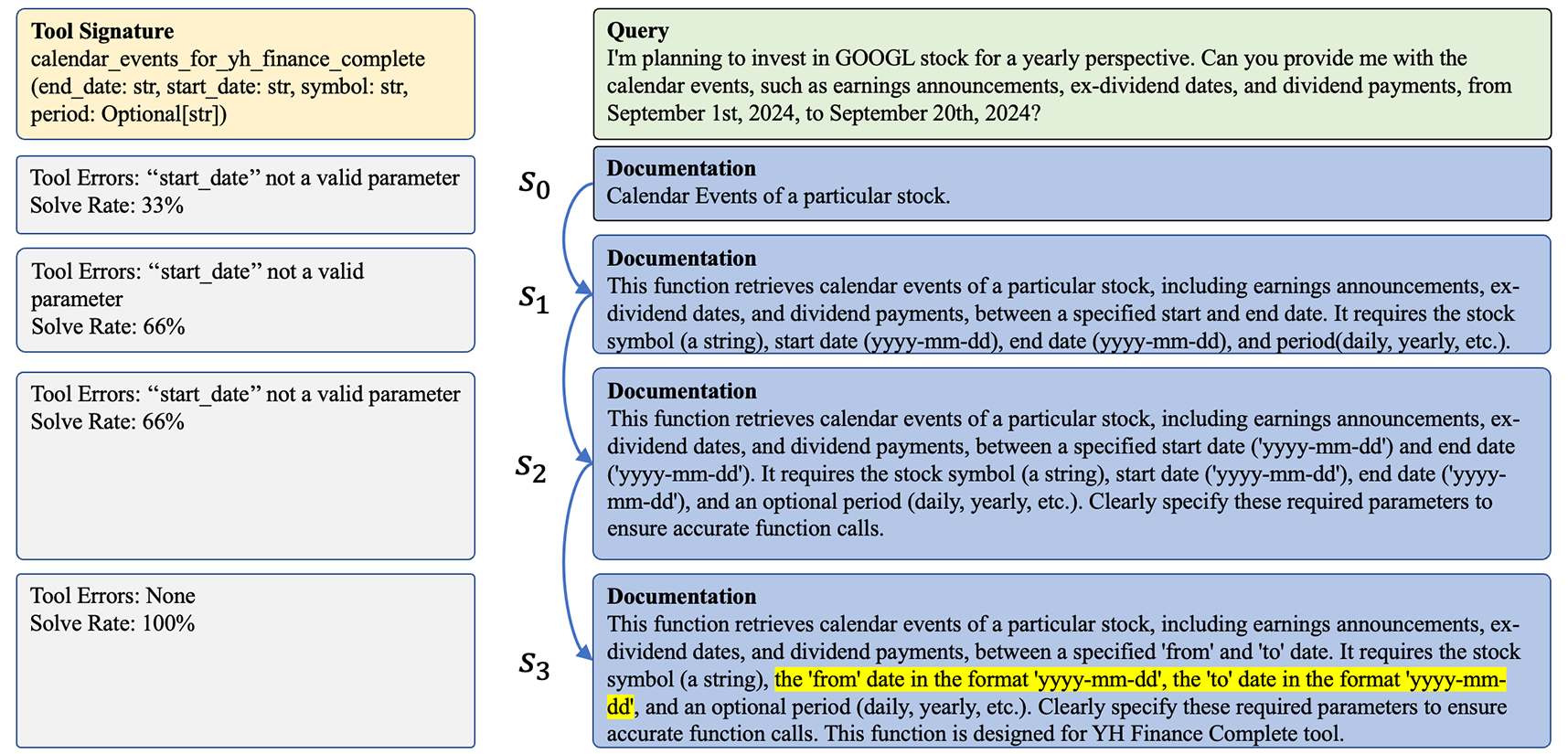}
    \caption{An example of \ours facing incorrect documentation. We show the beam search trajectory with the highest solve rate on the validation set. At each step, a new documentation is explored based on error feedback.}
    \label{fig:qualitative1}
    \vspace{-1em}
\end{figure*}

\paragraph{Ablation on Example Difficulty.}
We further examine how challenging examples influence performance by removing the \e{R_e} term during example generation and only using \e{R_q}. 
Without \e{R_e}, the model no longer targets difficult examples, yielding what we denote as \oursabbrev-Easy.
Under the same inference configuration, we substitute these examples in place of the original demonstrations and report results for LLaMA on StableToolBench in \tabref{tab:exp_hard}.
Ignoring difficult examples reduces final performance by an average of 1-2\% across all subsets, confirming that generating in-context examples with higher difficulty enhance tool usage.

\begin{table}[t!]
\setlength{\tabcolsep}{4pt}
\centering
\small
\begin{tabular}[t]{lcr}
\toprule
\bf Search strategy  & \bf \makecell{Pass\\Rate} & \bf \makecell{\# Proposals\\ Explored} \\
\midrule
ReAct & 64.3 & - \\
+\oursabbrev-MC (depth$=1$) & 65.2 & 3 \\
+\oursabbrev-MC (depth$=5$) & 66.7 & 15 \\
+\oursabbrev-MC (depth$=5$) \& \e{N_E}$=3$ & 68.9 & 45 \\
+\oursabbrev-Beam (depth$=5$) \& \e{N_E}$=3$ & \bf 70.7 & 135 \\
\bottomrule
\end{tabular}
\caption{Ablation on search strategies, on the I3-Inst subset of StableToolBench. LLaMA-70B is used for task model \e{\gM_T}. MC denotes Monte Carlo search.}
\label{tab:search}
\vspace{-1.7em}
\end{table}

\paragraph{Ablation on Search Strategy.}
We investigate alternative strategies for exploring the example-documentation space by comparing beam search in \ours to Monte Carlo search (MC) at various depths, sharing the same sample proposal scheme but with different number of proposals and sub-sampling approach.
We also study the role of \e{N_E} self-reflection steps during example generation.
Results are shown in \tabref{tab:search}, focusing on I3-Inst, which combines multiple tools from different categories and is thus more challenging.
The results show that MC underperforms beam search due to its limited exploration and is more subject to randomness, while deepening the search and adding self-reflection actions significantly improves performance, highlighting the importance of systematic search in optimizing examples and documentation.
Given the tradeoff between performance and efficiency, a suitable search method can be chosen when using \ours based on the user’s computational budget.
\paragraph{Qualitative Analysis.}
We give an example in \figref{fig:qualitative1} to show how \ours leverages tool play errors to refine tool documentation. 
The documentation is outdated, specifying \texttt{start\_date} and \texttt{end\_date} instead of \texttt{from} and \texttt{to}, and wrongly labeling them as required. 
Initially, \ours is confused by contradictory information but gradually incorporates more precise details, solves queries without needing those parameters, and eventually identifies the correct names, leading to superior performance. An additional example of typical improvements appears in appendix~\ref{appx:qualitative}.

\begingroup
\newlength{\cmrulespace}
\setlength{\cmrulespace}{\dimexpr
    \aboverulesep + \belowrulesep + \lightrulewidth\relax}
\newlength{\halfcmrulespace}
\setlength{\halfcmrulespace}{0.5\cmrulespace}
\begin{table*}[t!]
\setlength{\tabcolsep}{3.5pt}
\small
\begin{minipage}[t]{0.6\textwidth}
\centering
\begin{tabular}{lccccccccc}
\toprule
\bf Dataset & \multicolumn{3}{c}{\bf Completeness}& \multicolumn{3}{c}{\bf Conciseness}& \multicolumn{3}{c}{\bf Accuracy} \\
\midrule
\multirow{2}{*}{\raisebox{-\halfcmrulespace}{STB}} & \oursabbrev  & Raw & Equal& \oursabbrev  & Raw & Equal& \oursabbrev  & Raw & Equal \\
\cmidrule{2-10}
& \bf 85\% & 2\% & 13\% & 35\% & \bf 47\% & 18\% & \bf 43\% & 19\% & 37\% \\
\midrule
\multirow{2}{*}{\raisebox{-\halfcmrulespace}{BFCL}} & \oursabbrev  & Raw & Equal& \oursabbrev  & Raw & Equal& \oursabbrev  & Raw & Equal \\
\cmidrule{2-10}
& \bf 79\% & 5\% & 17\% & 15\% & \bf 70\% & 15\% & 18\% & 19\% & \bf 63\% \\
\bottomrule
\end{tabular}
\caption{Human evaluation on tool documentation quality, comparing completeness, conciseness, and accuracy of \ours versus raw (initial) documentation.}
\label{tab:eval_doc}
\end{minipage}
\hspace{0.02\textwidth}
\begin{minipage}[t]{0.37\textwidth}
\centering
\begin{tabular}{>{\raggedright\arraybackslash}m{0.35\textwidth}cc}
\toprule
\bf Examples & \bf Naturalness & \bf Alignment \\
\midrule
\multirow{2}{*}{\raisebox{-\halfcmrulespace}{BFCL Test Set}} & \multirow{2}{*}{\raisebox{-\halfcmrulespace}{4.39}} & \multirow{2}{*}{\raisebox{-\halfcmrulespace}{4.68}} \\
& & \\[\cmrulespace]
\midrule
\multirow{2}{*}{\raisebox{-\halfcmrulespace}{\ours}} & \multirow{2}{*}{\raisebox{-\halfcmrulespace}{4.17}} & \multirow{2}{*}{\raisebox{-\halfcmrulespace}{4.52}} \\
& & \\[\cmrulespace]
\bottomrule
\end{tabular}
\caption{Human evaluation on tool usage example (question-function call) quality, rated on a scale of 1 (low) to 5 (high).}
\label{tab:eval_example}
\end{minipage}
\vspace{-1em}
\end{table*}
\endgroup
\paragraph{Human Evaluation of Generation Quality.}
Since our primary objective is to improve LLMs’ tool-use capabilities, our main evaluation focuses on tool-use accuracy, which directly measures the effectiveness of our approach. 
While the quality of documentation and examples may not directly reflect tool-use improvements, they can provide complementary insights—particularly in enhancing clarity for human users and serving as secondary evidence of optimization quality. 
To assess these aspects, we conducted human evaluations on 50 randomly selected documentation outputs and 30 tool-use examples using six annotators. 
For documentation, we follow~\citet{quexploration} and assess completeness, conciseness, and accuracy relative to the original dataset-provided descriptions.
For tool-use examples, we adopt the evaluation criteria from~\citet{shen2024taskbench}, scoring the naturalness of the query and its alignment with the function call\footnote{We omit the complexity metric used in~\citet{shen2024taskbench}, as \ours generates single-tool usage examples.}, each on a scale of 1 to 5. 
As shown in \tabref{tab:eval_doc}, our optimized documentation achieves significantly higher completeness and comparable or better accuracy, though with reduced conciseness due to increased verbosity. 
As shown in \tabref{tab:eval_example}, the zero-shot-generated examples are rated similarly to manually curated ones in terms of both naturalness and alignment, demonstrating their practical utility.

\section{Related Work}
\paragraph{LLMs for Tool Use.}
Recent years have seen notable advances in employing large language models (LLMs) as agents to master tool use for solving complex tasks~\citep{mialon2023augmented,qin2024toollearningfoundationmodels}, thereby enhancing LLMs' capabilities in multi-modal understanding~\citep{gupta2023visual, suris2023vipergpt, wu2023visual}, programming tools~\citep{gao2023pal, paranjape2023art,team2023gemini, zhang2023natural,cai2024large}, and other domain-specific functionalities.
The conventional approach involves training base models with tool-use data~\citep{thoppilan2022lamda,dubey2024llama} or fine-tuning LLMs~\citep{patil2023gorilla, schick2023toolformer, yang2024gpt4tools,parisi2022talm,liu2025toolace,lin2025robust}, but may require continual learning as new tools are added. 
\citet{hao2024toolkengpt} addressed scalability by training tool embeddings plug-and-play usage, though still requiring labeled data. 
Alternatively, LLMs can be augmented with meta-prompts or tool-use instructions at inference time~\citep{lu2023chameleon,shen2023hugginggpt, song2023restgpt, qin2024toolllm, zhuang2024toolchain}. 
As the range of applications and tools expands, enhancing LLMs' capacity to handle new tools remain pivotal, which we address through \ours.

\paragraph{Tool-Use Instructions and Optimization.}
Tool documentation and example demonstrations are key to prompting LLMs for effective tool use. 
\citet{hsieh2023tool} highlighted the risk of hallucinations when documentation is lacking, and \citet{xu2023tool} showed performance declines without in-context examples. 
To automate generating tool-use instances, \citet{shen2024taskbench} leveraged a graph of tool relations for back-instructing queries, relying on the availability of external tool graphs. 
\citet{yuan2024easytool} proposed direct prompting to rewrite tool documentation, which relies on labeled documentation examples and lacks systematic optimization and the ability to measure optimization quality. 
\citet{quexploration} explored LLMs interacting with tools, focusing on tool documentation only with single-thread iterative refinement.
Automatic prompt tuning~\citep{pryzant-etal-2023-automatic, wang2024promptagent} adapts prompts to domain-specific tasks but require held-out test sets, rendering it unsuitable for zero-shot tool instruction rewriting~\citep{wu2024avatar}. 
These constraints underscore the need for approaches that optimize tool instructions and demonstrations without labeled data or manual effort, which \ours achieves by interacting directly with the tool itself.

\section{Conclusion}
We present \ours, an automated framework that iteratively refines tool documentation and creates usage demonstrations, enhancing LLMs' tool use in zero-shot settings. 
Through a search-based trial-and-error process with self-reflection, \ours allows models to explore and improve tool use without labeled data or extensive manual effort. 
This approach addresses the limitations of methods that rely on handcrafted prompts or labeled data, providing a scalable and task-agnostic solution for real-world applications. 

\section*{Acknowledgments}
This research is supported by the MIT-IBM Watson AI Lab and the Centre for Perceptual and Interactive Intelligence (CPII) Ltd. under the Innovation and Technology Commission’s InnoHK Scheme.
The views and conclusions are those of the authors and should not be interpreted as representing those of IBM.

\section*{Limitations}
In this work, \ours optimizes a single tool, but it's still applicable to queries that require multiple tools by optimizing each tool independently, and then using the optimized examples and documentation from multiple tools together at inference.
Although results (on Multiple and Multiple-Parallel in BFCL and the entirety of StableToolBench) on multi-tool queries show sizeable performance gains,
scaling the optimization process itself from single-tool scenarios to multiple tools can likely enhance LLM's tool use effectiveness even more.
Additionally, for example demonstrations, we use rejection sampling to generate tool invocations first, which do not work for functions whose parameter space is too large, for instance parameters that take long ID string inputs or authentication tokens that require calls to other tools beforehand.
Exploring multi-tool dependencies could potentially resolve this issue and improve tool play.
In our work we focus on tool documentation and demonstrations only, relegating other information as meta-information, which could be potential next steps to explore.

\section*{Ethics Statement}
We used publicly available models and datasets for training and evaluation, and did not collect data or any personal information in this work. The trained models could however potentially be misused and pose ethical risks typical of large language models when deployed in real-world applications, if not thoroughly audited.

\bibliography{anthology,tool}

\appendix
\section{Dataset License}
BFCL and StableToolBench are both licensed under Apache License 2.0. We adhere to intended uses stated in the license.

\section{Additional Implementation Details}\label{appx:inference}
For task LLM \e{\gM_T}, \texttt{llama-3.1-8b-instruct} is used for LLaMA-8B, \texttt{llama-3.3-70b-instruct} for LLaMA-70B, \texttt{gpt-3.5-turbo-0125} for GPT-3.5, \texttt{gpt-4o-2024-11-20} for GPT-4o, \texttt{toolace-2-llama-3.1-8b} for ToolACE-8B, and \texttt{hammer2.1-7b} for Hammer-7B.
On StableToolBench, we use \texttt{llama-3.3-70b-instruct} as the evaluation LLM, which produces stable assessments.

For GPT models, we keep the exact same inference setting as used in the original benchmark, except for requiring it to return a single action at each step from the provided toolset by supplying a flag to the OpenAI API.
For LLaMA models, on BFCL we follow the exact official inference prompts, and on StableToolBench we adapt the ReAct prompts into LLaMA-3's prompt format but keep everything else fixed as much as possible.
For ToolACE and Hammer on BFCL, we also follow the exact official inference prompts. 

During inference, for \ours we set the number of tool-use example demonstrations to 5 per tool for BFCL and 1 per tool for StableToolBench, as StableToolBench averages more tools per query. The sampling temperature is set to $0.001$. We report scores averaged over 3 independent runs.

As both our optimization and inference stages perform only LLM inference, we call hosted inference APIs and do not report total computation in GPU hours.
An estimated 1M API calls were made in total for this work.

\begingroup
\setlength{\tabcolsep}{4pt}
\begin{table*}[t!]
\small
\centering
\begin{tabular}{lL{3.4cm}C{1.2cm}C{1.2cm}C{1.2cm}C{1.2cm}C{1.2cm}>{\columncolor[gray]{0.9}}C{1.2cm}>{\columncolor[gray]{0.9}}C{1.2cm}}
\toprule
\bf Base Model & \bf Method  & \bf Simple-Python & \bf Simple-REST & \bf Multiple & \bf Parallel & \bf Multiple Parallel & \bf Weighted Avg & \bf Avg \\
\midrule
\multirow{2}{*}{LLaMA-8B} & Prompting & 91.0 & 67.1 & 92.0 & 84.0 & 70.0 & 81.3 &  80.8 \\
 &+\ours & 94.0 & 81.4 & 92.0 & 82.0 & 82.5 & \bf 86.1 & \bf 86.4 \\
\midrule
\multirow{2}{*}{LLaMA-70B} & Prompting & 96.0 & 90.0 & 92.0 & 80.0 & 72.5 & 84.4 & 86.1 \\
&+\ours  & 95.0 & 88.6 & 92.0  & 84.0 & 87.5 & \bf 88.8 & \bf 89.4  \\
\midrule
\multirow{2}{*}{GPT-3.5} & Function-calling & 92.0 & 87.1 & 84.0 & 76.0 & 50.0 & 74.9 & 77.8 \\
&+\ours  & 98.0 & 88.6 & 86.0 & 84.0 & 80.0 & \bf 85.8 & \bf 87.3 \\
\midrule
\multirow{2}{*}{GPT-4o} & Function-calling & 93.0 & 95.7 & 92.0 & 88.0 & 75.0 & 87.3 & 88.7 \\
&+\ours  & 98.0 & 94.3 & 94.0 & 88.0 & 82.5 & \bf 90.3 & \bf 91.6 \\
\midrule
\multirow{2}{*}{ToolACE-8B} & Function-calling & 92.0 & 84.3 & 86.0 & 80.0 & 57.5 & 77.9 & 80.0 \\
&+\ours  & 93.0 & 84.3 & 90.0 & 82.0 & 72.5 & \bf 83.3 & \bf 84.4 \\
\midrule
\multirow{2}{*}{Hammer-7B} & Function-calling & 93.0 & 71.4 & 78.0 & 82.0 & 62.5 & 76.2 & 77.4 \\
&+\ours  & 92.0 & 82.9 & 86.0 & 82.0 & 75.0 & \bf 82.6 & \bf 83.6 \\
\bottomrule
\end{tabular}
\caption{Results on BFCL Executable, with parameter description dropout $p=0.5$. Accuracy scores are shown.}
\vspace{-1em}
\label{tab:bfcl_noisy}
\end{table*}
\endgroup
\begingroup
\setlength{\tabcolsep}{4pt}
\begin{table*}[t!]
\small
\centering
\begin{tabular}{lL{3.4cm}C{1.2cm}C{1.2cm}C{1.2cm}C{1.2cm}C{1.2cm}>{\columncolor[gray]{0.9}}C{1.2cm}>{\columncolor[gray]{0.9}}C{1.2cm}}
\toprule
\bf Base Model & \bf Method  & \bf Simple-Python & \bf Simple-REST & \bf Multiple & \bf Parallel & \bf Multiple Parallel & \bf Weighted Avg & \bf Avg \\
\midrule
\multirow{2}{*}{LLaMA-8B} & Prompting & 85.0 & 62.9 & 84.0 & 82.0 & 55.0 & 73.7 & 73.8  \\
 &+\ours & 88.0 & 78.6 & 90.0 & 82.0 & 67.5 & \bf 80.7 & \bf 81.2 \\
\midrule
\multirow{2}{*}{LLaMA-70B} & Prompting & 86.0 & 87.1 & 86.0 & 76.0 & 55.0 & 75.9 & 78.0 \\
&+\ours  & 89.0 & 87.1 & 92.0 & 80.0 & 65.0 & \bf 81.3 & \bf 82.6  \\
\midrule
\multirow{2}{*}{GPT-3.5} & Function-calling & 84.0 & 87.1 & 78.0 & 72.0 & 40.0 & 68.9 & 72.2 \\
&+\ours  & 92.0 & 84.3 & 82.0 & 86.0 & 52.5 & \bf 77.2 & \bf 79.4 \\
\midrule
\multirow{2}{*}{GPT-4o} & Function-calling & 85.0 & 95.7 & 84.0 & 84.0 & 47.5 & 76.5 & 79.2 \\
&+\ours  & 91.0 & 94.3 & 92.0 & 88.0 & 60.0 & \bf 83.2 & \bf 85.0  \\
\midrule
\multirow{2}{*}{ToolACE-8B} & Function-calling & 88.0 & 80.0 & 86.0& 82.0 & 52.5 & 76.1 & 77.7 \\
&+\ours  & 89.0 & 85.7 & 90.0 & 86.0 & 65.0 & \bf 82.1 &  \bf 83.1 \\
\midrule
\multirow{2}{*}{Hammer-7B} & Function-calling & 85.0 & 75.7 & 76.0 & 84.0 & 52.5 & 73.2 & 74.6 \\
&+\ours  & 90.0 & 82.9 & 88.0 & 86.0 & 70.0 & \bf 82.6 & \bf 83.4 \\
\bottomrule
\end{tabular}
\caption{Results on BFCL Executable, with parameter description dropout $p=1.0$. Accuracy scores are shown.}
\vspace{-1em}
\label{tab:bfcl_noisy_p0}
\end{table*}
\endgroup

\section{Details on Robustness Experiments}\label{appx:robustness}
To assess the robustness of \ours, we artificially introduce noise to tool documentation in BFCL Executable by dropping each parameter description with a probability $p$, leaving general tool descriptions and signatures intact.
As shown in \tabref{tab:bfcl_noisy}, with $p=0.5$, this degradation reduces baseline performance by about 5\% for the LLaMA models and GPT-4o, and by 10\% for GPT-3.5, especially for the most challenging Multiple-Parallel category (c.f. \tabref{tab:bfcl_main}).
With \ours, performance consistently exceeds baselines across all models, improving by 3-4\% for the larger LLaMA-70B and GPT-4o models, 5\% for LLaMA-8B, and most significantly, recovering the full 10\% loss for GPT-3.5.

In \tabref{tab:bfcl_noisy_p0}, we further include experiments on BFCL with an even higher parameter description dropout with $p=1.0$, that is, dropping out all parameter descriptions. 
The documentation still contains information from the overall tool description and parameter name in this scenario.
These results illustrate \ours's robustness to incomplete documentation, and especially shines when initial documentation is poor.

\section{Additional Qualitative Example}\label{appx:qualitative}
\begin{figure*}[ht!]
    \centering
    \includegraphics[width=\linewidth]{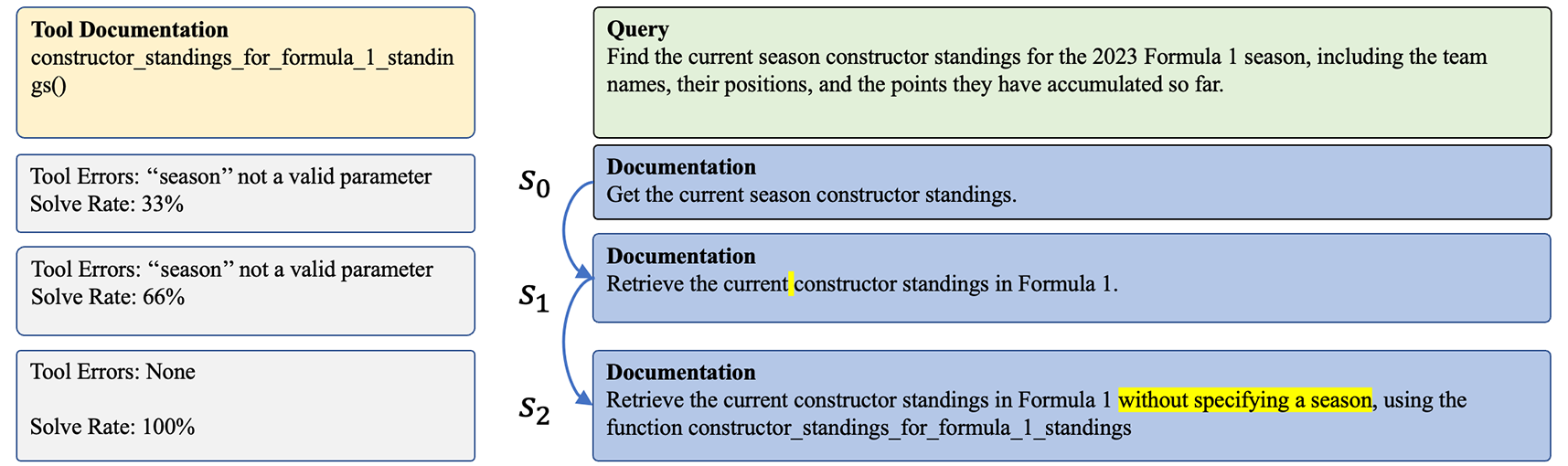}
    \caption{A typical example of \ours assisting LLMs in correcting errors in generating parameter values. The task model \e{\gM_T} (LLaMA-8B) gets confused by the query specifying the year, which \ours first attempts to remove ``year'' from the description, and further explicitly prompts \e{\gM_T} to not use the parameter.}
    \label{fig:qualitative2}
\end{figure*}

We show another example to illustrate how \ours commonly aids LLMs' tool usage in \figref{fig:qualitative2}.

\begin{algorithm*}[t!]
\caption{\sc ExampleOptimizationStep}
\label{alg:ex}
\begin{algorithmic}[1]
\Require $\gS_t=E_t=(x_t, F, I_t, y_t)$: tool-use example, $\gD_0$: documentation, $a_t$: reflection
\Ensure $\gS_{t+1}=E_{t+1}=(x_{t+1},F,I_{t+1},y_{t+1})$: updated example, $r_{t+1}$: score, $a_{t+1}$: reflection
\State $c\gets$ false
\While{$\neg c$} \Comment{\small Rejection Sampling of tool invocation $i$}
    \State $I_{t+1}\sim p_{\gG_E}(I|I_t, c, \gD_0, a_t, m_1)$ \Comment{\small Sample candidate tool invocation, incorporating reflection $a_t$}
    \State $o_{t+1}\gets F(I_{t+1})$ \Comment{\small Execute tool function}
    \State $c\sim p_{\gG_E}(c|I_{t+1}, o_{t+1}, \gD_0, m_2)$ \Comment{\small Verify validity of tool call}
\EndWhile
\For{$n\gets 1$ to $N_E$} \Comment{\small Rollout for with self-reflection policy}
    \State $x_{t+1}\sim p_{\gG_E}(x|I_{t+1},o_{t+1},\gD_0,m_3)$ \Comment{\small Sample user query $x$}
    \State $y_{t+1}\sim p_{\gG_E}(y|I_{t+1},o_{t+1},x_{t+1},\gD_0, m_4)$ \Comment{\small Sample corresponding answer $y$}
    \State \e{\gR_q}$\sim p_{\gG_E}(r|y_{t+1},I_{t+1},o_{t+1},x_{t+1},\gD_0, m_5)$ \Comment{\small Evaluate example quality}
    \State \e{\gR_e}$\gets -\gP\{\gM_T(x_{t+1};\gD_0, \varnothing);y_{t+1}, F, I_{t+1}\}$ \Comment{\small Evaluate example difficulty}
    \State $r_{t+1}\gets$\e{\gR_q+\gR_e}
    \State $a_{t+1}\sim p_{\gG_E}(a|r_{t+1},y_{t+1},i_{t+1},o_{t+1},x_{t+1},\gD_0, m_6)$ \Comment{\small Generate self-reflection action}
\EndFor
\end{algorithmic}
\end{algorithm*}
\begin{algorithm*}[t!]
\caption{\sc DocumentationOptimizationStep}
\label{alg:desc}
\begin{algorithmic}[1]
\Require $\gS_t=D_t$: documentation, $\gE=\{(x_j,F,I_j,y_j)\}_{j=1}^W$: validation set, $a_t$: reflection
\Ensure: $\gS_{t+1}=\gD_{t+1}$: updated documentation, $r_{t+1}$: score, $a_{t+1}$: reflection
\State $\gD_{t+1}\sim p_{\gG_D}(\gD|\gD_t,a_t,m_7)$ \Comment{\small Sample documentation $\gD_{t+1}$ from $\gG_D$, applying reflection $a_t$}
\State $\hat{M}_j, e_j\gets \gM_T(x_j;\gD_{t+1},\varnothing)~\forall j$ \Comment{\small Gather I/O \& errors $e_j$ from running $\gM_T$ on validation set $\gE$ with $\gD_{t+1}$}
\State $r_{t+1}\gets \E_j [\gP\{\hat{M}_j;y_j,F,I_j\}]$ \Comment{\small Evaluation score on $\gE$}
\State $a_{t+1}\sim p_{\gG_D}(a|\gD_{t+1},r_{t+1},\{x_j,I_j,y_j,\hat{M}_j,e_j\},m_8)$ \Comment{\small Self-reflection action}
\end{algorithmic}
\end{algorithm*}
\section{Detailed Optimization Procedures}\label{appx:algo}
Below, we present the full procedures for both example demonstration optimization and documentation optimization as pseudo-code, shown in \Algref{alg:ex} and \Algref{alg:desc}, respectively.
Please refer to \Secref{sec:notation} for notations.
In the algorithms, $m_i$ represents meta-prompts.

\section{Meta-Prompts}\label{appx:prompts}
The meta-prompts are listed below in tables~\ref{tab:prompt_m1} through~\ref{tab:prompt_m8}.
\begin{table*}[h]
\small
\centering
\begin{tabular}{L{\linewidth}}
\toprule
 You are given an API tool with the following documentation: \{{\tt Documentation}\} \\\\
Your task is to write 1 example API call for the given API tool given its purpose and parameters list. The API call you produced will be executed as function call later and return result if correct, or error if you provide incorrect syntax, format, or parameters. Given the documentation and description, think of possible example API calls and produce those that are likely to be correctly executed. Think of parameter values that are reasonable, make sense, and are likely API calls that people use in the real world. The generated API call must be executable and real. Parameter values must be filled in and not placeholding text. You must include the required parameters, and optionally give parameters that are labeled as "optional parameters". Do not hallucinate and produce parameters that are not under "required" or "optional". Produce diverse parameter values, but be factual and do not use fake parameters. 
\\
You can only use the given function \{{\tt function\_name}\}. Create an API call that include the function name, and the parameters to be input to the API. Include all the required and optional parameters in a single dictionary without separating them. Do not include the URL or other irrelevant information. The output should be in the following JSON format that represents a function call:
\{"name": "function\_name","parameters": \{"properties": \{"parameter\_1": value 1\}\}\}
You must strictly follow the output format, including "name", "parameters", "properties", and parameters.
\\\\
Previously you generated the following API calls for this function, which where then executed and critiqued:\\
fn\_call="\{{\tt fn\_call}\}" fn\_output="\{{\tt fn\_output}\}" status=\{{\tt status}\} reflection="\{{\tt reflection}\}"

         \\
        \bottomrule
    \end{tabular}
    \caption{Meta-prompt $m_1$}
    \label{tab:prompt_m1}
\end{table*}

\begin{table*}[h]
\small
\centering
\begin{tabular}{L{\linewidth}}
\toprule
 You are given an API tool with the following documentation: \{{\tt Documentation}\} \\\\
 Previously you were asked to write an example API call for the function \{{\tt function\_name}\} given its purpose and parameters list, and you generated the following function call: \{{\tt fn\_call}\}. The function call you produced was later executed and returned the following result: "\{{\tt fn\_output}\}". 
 \\\\
 Your task is to analyze the response and check if there are any errors. \\
1. If there are no errors and everything looks reasonable, give an err\_code of 0, and don't provide analysis.\\
2. If there is an error, give an err\_code of -1. Then in your analysis, describe and analyze in detail why the error occurred based on the error message. Then, based on your analysis, give detailed suggestions to improve the function call so that no errors will be produced. You must give detailed analysis and suggestions, do not simply repeat the error message. The analysis and suggestions should be in the "analysis" field in the output.\\
Note that even if the "error" field in the result is empty, the "response" field may contain an error when using the function call. If this is the case you must treat this as an error and analyze the failure. The response field may also be in HTML format.  
\\\\
Your output should be in the following JSON format:
\{"analysis": your analysis and suggestions, "err\_code": error code (-1 for error, 0 for correct)\}\\
\bottomrule
\end{tabular}
\caption{Meta-prompt $m_2$}
\label{tab:prompt_m2}
\end{table*}

\begin{table*}[h]
\small
\centering
\begin{tabular}{L{\linewidth}}
    \toprule
 You are given an API tool with the following documentation: \{{\tt Documentation}\} \\\\
For the function \{{\tt function\_name}\}, you are given the following function call: \{{\tt fn\_call}\}, and executing the function call returned the following result: \{{\tt fn\_output}\}. 
\\\\
Your task is to generate a user instruction in natural language that requires the given function call to be completed. Here are some guidelines to follow:\\
1. The instruction must be a scenario or problem that cannot be solved without calling the given function \{{\tt function\_name}\}. This is your main objective.\\
2. You should not directly or explicitly ask for the function to be called; the problem itself must inherently be solved by the function.\\
3. Based on the function, function call, its parameters, parameter values, and function execution responses, you should produce a real and reasonable instruction. \\
4. You must use information from the parameter values of the function call to create the response. You must include the value of every parameter from the given function call in the user instruction you generated, including each list/dict element of the parameter values. Do not ignore any parameters/values from the function call.\\
5. You must not include specific function calls in your response. You should not explicitly show the function names. You should also never explicitly name the parameter names in your response. You should not show any variable names.\\
6. Your response has to be in natural language. Do not show any variables, function calls, or code. \\
7. You should respond in the user's first-person perspective.\\
8. You are a human user. You are asking a question or giving an instruction. Do not answer in the perspective of an AI assistant. Remember, the user does not know about the API function and thus cannot ask to call the function.\\
9. Remember, you are asking a question, so do not answer your own question in the response. Your goal is to give a querying instruction or question, not producing answers or function calls.\\\\
Your output should be in the following JSON format:
\{"instruction": generated instruction\}
\\\\
Previously you generated the following instructions for this function call, which were rated and analyzed:
instruction="\{{\tt instruction}\}" score=\{{\tt score}\}

Based on these ratings, you are given the following analysis: \{{\tt reflection}\}. You should improve your instructions based on these suggestions.\\
    \bottomrule
\end{tabular}
\caption{Meta-prompt $m_3$}
\label{tab:prompt_m3}
\end{table*}

\begin{table*}[h]
\small
\centering
\begin{tabular}{L{\linewidth}}
    \toprule
 You are given an API tool with the following documentation: \{{\tt Documentation}\} \\\\
You are given the following instruction: "\{{\tt instruction}\}"
To produce a response to the instruction, you made an API call to the given tool, which returned the following results: \{{\tt fn\_output}\}

Given the instruction and the results of API call, produce an effective and short answer to the user in natural language. Your answer must be based on the results of the API call, do not hallucinate or answer anything not in the API results. You must not include code, comments, JSON data structures, notes, or other irrelevant information in your answer. If there is an error or failure using the tool, you must report the error in your answer and do not make things up, especially when you receive an input about invalid parameters. 
      \\
    \bottomrule
\end{tabular}
\caption{Meta-prompt $m_4$}
\label{tab:prompt_m4}
\end{table*}

\begin{table*}[h]
\small
    \centering
    \begin{tabular}{L{\linewidth}}
        \toprule
You are given an instruction "\{{\tt instruction}\}", function call "\{{\tt fn\_call}\}" and an answer "\{{\tt answer}\}", your task is to give a `score` based on the following rules:\\
1. You must return 1 if any of the following conditions is met (for instruction only): (1) instruction is empty, nonsense, or not in natural language; or (2) instruction is explicitly including function calls or asking for function calls or contains function names; or (3) instruction includes exact function parameter names; or (4) instruction includes code or variable assignment; or (5) instruction is longer than 3 sentences or 300 letters; or (6) instruction does not include a question, query, request, or problem to be solved; or (7) instruction is not in first-person perspective, or is in the perspective of an AI assistant instead of a user; or (8) any parameter value in the function call is not present in the instruction\\
An instruction that satisfies any of these conditions is a bad instruction and should be scored a 1.\\
2. If the answer is a error message or mentions any errors (API error, invalid parameter error, ..., etc.), mentions cannot use API or cannot respond, return 1.\\
3. If the answer is a positive response for the given instruction, you have to further check.\\
3.1 If the answer is not sufficient to determine whether they solve the instruction or not, return 2.\\
3.2 If you are confident that the answer is sufficient to determine whether the solve the instruction or not, return 3 if solvable or 1 if unsolvable.
\\\\
Finally, organize your output in the following JSON format:\{"analysis": your reasoning, "score": score\}\\
        \bottomrule
    \end{tabular}
\caption{Meta-prompt $m_5$}
    \label{tab:prompt_m5}
\end{table*}
\begin{table*}[h]
\small
    \centering
    \begin{tabular}{L{\linewidth}}
        \toprule
 You are given an API tool with the following documentation: \{{\tt Documentation}\} \\\\
Previously, given the function call \{{\tt fn\_call}\}, you were asked to generate example instructions that require the use of the function \{{\tt function\_name}\} to complete. The example instructions generated by you were then scored by an expert on whether the instructions can be fulfilled using the given API function. Scores are in a scale between 1 (lowest) and 3 (highest). 
\\\\
Below are the generated instructions, scores, and analyses:\\
instruction="\{{\tt instruction}\}" score=\{{\tt score}\} analysis="\{{\tt analysis}\}"
\\\\
Task:\\
1. Firstly, identify and contrast the patterns of instructions and function calls that have achieved good scores with those that have not. If there are no bad scores, only summarize the patterns of the good ones.\\
2. Next, specify the suggestions that can lead to improved performance for the generated instructions and function calls with bad scores. You should focus on capturing the high-level pattern of the examples relevant to the API documentation. Note that both the function and the function call cannot be changed, and focus your suggestions on how to improve the example instructions, including deciding what information to use from parameters of the function call.\\
\bottomrule
\end{tabular}
\caption{Meta-prompt $m_6$}
    \label{tab:prompt_m6}
\end{table*}

\begin{table*}[h]
\small
    \centering
    \begin{tabular}{L{\linewidth}}
        \toprule
 You are given an API tool with the following documentation: \{{\tt Documentation}\} \\\\
Previously, the given tool was used in solving instructions by a tool assistant with the following descriptions:\\
Iteration \#\{{\tt iteration}\}, description="\{{\tt description}\}", score=\{{\tt score}\}\%, stdev=\{{\tt stdev}\}.
\\\\
Furthermore, an analysis was performed on the descriptions for the previous iterations: "\{{\tt analysis}\}"\\\\
Your task is to further enhance the description for the function \{{\tt function\_name}\} based on these results for the next iteration, with the objective of maximizing the score, minimizing the stdev, and help the assistant correctly use the function without errors. The descriptions for each parameter might be unclear, underspecified, or incorrect, so you should include clear parameter descriptions and usage for every single required and optional parameter, including its type, usage, and possible values. Be as clear, descriptive, and comprehensive as possible. Be factual and do not consider parameters that are not listed. Incorporate the analysis and generate the enhanced descriptions. \\
\bottomrule
    \end{tabular}
\caption{Meta-prompt $m_7$}
    \label{tab:prompt_m7}
\end{table*}

\begin{table*}[h]
\small
 \centering
 \begin{tabular}{L{\linewidth}}
 \toprule
 You are given an API tool with the following documentation: \{{\tt Documentation}\} \\\\
Previously, the given tool was used in solving instructions by a tool assistant with the following descriptions:\\
Iteration \#\{{\tt iteration}\}, description="\{{\tt description}\}"\\\\
Here are the instructions the assistant tried to solve with this tool description, with their corresponding answers and errors produced by the assistant:\\
instruction="\{{\tt instruction}\}", answer="\{{\tt answer}\}", errors: function\_call=\{{\tt function\_name}\}, arguments=\{{\tt arguments}\}, error=\{{\tt error\_message}\}, ground truth should be \{{\tt fn\_call}\}\\\\
Overall the performance of this description is: score=\{{\tt score}\}\\\\
Now your task is to critique the descriptions based on these results. A good description maximizes the score, minimizing the stdev, and helps the assistant correctly use the function without errors. In your analysis:\\
(1) Identify how the descriptions affect the function call errors of the assistant. Be specific on which errors the assistant tends to make, and find patterns in the description that causes the assistant to make such errors.\\
(2) Identify and contrast the patterns of descriptions that have achieved good scores (> 60\%) with those that have not. Analyze how the description can be improved. \\
\bottomrule
\end{tabular}
\caption{Meta-prompt $m_8$}
    \label{tab:prompt_m8}
\end{table*}

\end{document}